\documentclass{article} 
\usepackage{iclr2023_conference_tinypaper,times}

\usepackage{amsmath,amsfonts,bm}

\def\eqref#1{equation~\ref{#1}}

\def\1{\bm{1}}

\DeclareMathAlphabet{\mathsfit}{\encodingdefault}{\sfdefault}{m}{sl}
\SetMathAlphabet{\mathsfit}{bold}{\encodingdefault}{\sfdefault}{bx}{n}

\usepackage{hyperref}
\usepackage{url}
\usepackage{graphicx}
\usepackage{multirow}
\usepackage{caption}
\usepackage{subcaption}

\usepackage{booktabs}
\usepackage{nicefrac}
\usepackage{microtype}
\usepackage{xcolor}
\usepackage{multirow}
\usepackage{epsfig}
\usepackage{amsmath}
\usepackage{diagbox}
\usepackage{float}
\usepackage{pifont}
\usepackage{color}
\usepackage{xspace}
\usepackage{caption}
\usepackage{wrapfig}
\usepackage{varwidth}
\newcolumntype{M}{>{\begin{varwidth}{4cm}}l<{\end{varwidth}}}

\title{Language Model Knowledge Distillation for Efficient Question Answering in Spanish}

\author{Adrián Bazaga, Pietro Liò \& Gos Micklem \\
University of Cambridge\\
\texttt{\{ar989,pl219,gm263\}@cam.ac.uk}
}

\iclrfinalcopy 
\begin{document}

\maketitle

\begin{abstract}
Recent advances in the development of pre-trained Spanish language models has led to significant progress in many Natural Language Processing (NLP) tasks, such as question answering. However, the lack of efficient models imposes a barrier for the adoption of such models in resource-constrained environments. Therefore, smaller distilled models for the Spanish language could be proven to be highly scalable and facilitate their further adoption on a variety of tasks and scenarios. In this work, we take one step in this direction by developing SpanishTinyRoBERTa, a compressed language model based on RoBERTa for efficient question answering in Spanish. To achieve this, we employ knowledge distillation from a large model onto a lighter model that allows for a wider implementation, even in areas with limited computational resources, whilst attaining negligible performance sacrifice. Our experiments show that the dense distilled model can still preserve the performance of its larger counterpart, while significantly increasing inference speedup. This work serves as a starting point for further research and investigation of model compression efforts for Spanish language models across various NLP tasks.
\end{abstract}

\section{Introduction}

In the last years, recent advancements in the field of NLP have panned the way for progress on a variety of downstream tasks, primarily by fine-tuning pre-trained language models (PLMs) \citep{vaswani2017, devlin2019bert, liu2019roberta, yang2019} on large-scale task-specific datasets. These models are typically trained on large and high-quality annotated corpora, which are are usually scarce for languages other than English, posing a significant disadvantage for progressing multilingual NLP research. Therefore, resources for languages such as Spanish, the fourth most spoken language, remain underrepresented in terms of amount of training data available. Moreover, despite the fact that some language models are available for the Spanish language \citep{gutierrezfandino2022, CaneteCFP2020, delarosa2022bertin, perez-etal-2022-robertuito}, operating them in resource-constrained environments remains an under-explored challenge. To overcome this limitation, we propose employing knowledge distillation \citep{jiao-etal-2020-tinybert, boreshban2021improving} to transfer the intrinsic knowledge of a large Spanish RoBERTa language model into a much compact and efficient model, named as SpanishTinyRoBERTa, while maintaining its performance. In particular, we focus on the question answering task \citep{rajpurkar2016squad, carrino2019automatic}, where the goal of the model is to provide an answer given a context and a question over that context. Our experiments show that SpanishTinyRoBERTa attains negligible performance reduction on the question answering task with respect to the teacher model, whilst increasing significantly inference speed and reducing resource requirements. The ultimate goal of our work is to facilitate efforts in the development of efficient language models for the Spanish language, thereby breaking down computational barriers and promoting adoption of NLP technologies.

\section{Methodology} \label{sec:methodology}

We propose developing a Spanish TinyRoBERTa model using the knowledge distilled from a Spanish RoBERTa-large model in the SQuAD-es QA task \citep{carrino2019automatic}. In the context of language models, knowledge distillation can be modeled as penalizing the difference of feature representations between the teacher and the model, therefore aiming to minimize the following objective function:

\begin{align}
\label{eq:generic_kd}
\mathcal{L}_{\text{KD}} =  \sum_{x \in \mathcal{M}} L\big(f^S(x), f^T(x)\big),
\end{align}

where $L(\cdot)$ is a loss function that computes the discrepancy between teacher and student models, $x$ is the text input and $\mathcal{M}$ denotes the training dataset. In this work, we employ the knowledge distillation technique introduced in TinyBERT \citep{jiao-etal-2020-tinybert} to reproduce the behavior of the larger model and leveraging the knowledge transferred from it. During model training, as the student model (SpanishTinyRoBERTa) contains much less layers than the teacher, we map layers in the student to the teacher by using a layer mapping function, so that the student learns from intermittent layers of the teacher model (more details in Appendix \ref{appendix:distillation_process}).

\section{Experimental Setup and Results}

To demonstrate the effectiveness of our distillation for Spanish question answering (QA) tasks, we use the SQuAD-es dataset \citep{carrino2019automatic}, which is a Spanish version of SQuAD \citep{rajpurkar2016squad}. To address questions answering as a learning problem, we treat the QA task as the problem of sequence labeling which predicts the possibility of each token as the start or end of answer span. We instantiate a tiny RoBERTa student model, with number of layers $M$=6, the hidden size as $d_{h}$=512, the feedforward size as $d_{ff}$=3072 and the head number as $h$=16), account for a total of 51.4M parameters. If not specified, this student model is referred to as the SpanishTinyRoBERTa. We use as teacher model a Spanish RoBERTa-large model pre-trained on a corpus from the National Library of Spain \citep{gutierrezfandino2022}. The teacher model has layers $N$=12, a hidden size $d_{h}$=1024, feed-forward size $d_{ff}$=4096 and $h$=16 heads, accounting for a total of 355M parameters. We use four different models as baselines: the Multilingual BERT (mBERT), a Spanish BERT-base \citep{CaneteCFP2020}, a Spanish RoBERTa-base (124M parameters) and a Spanish RoBERTa-large as teacher model. As the main focus of this work is to improve running time efficiency, we use these baseline models to measure both the inference speed and performance when compare with our lighter model. More details on the training hyperparameters in Appendix \ref{appendix:training_details}.

\begin{table}[H]
    \renewcommand\arraystretch{1.}
    \centering
    \small
\begin{tabular}{lccc}
\toprule
Model    & F1 (\%) & EM (\%) & Inference Speedup\\
\hline
Spanish RoBERTa-large (teacher)  & 87.50 & 78.30 & 1.0x      \\ 
Multilingual BERT (mBERT) & 77.60 & 61.80 & 3.0x \\ 
Spanish BERT-base & 82.15 & 73.59 & 2.4x \\ 
Spanish RoBERTa-base & 81.80 & 72.30 & 2.5x   \\ 
\hline
SpanishTinyRoBERTa (ours) & 80.52 & 71.23 & 4.2x  \\ 
\hline
\end{tabular}
\caption{Comparison of the different models on the SQuAD-es Spanish Question Answering task using the F1 and Exact Match (EM) metrics.}
  \label{table:task_results}
  \vspace{-.2in}
\end{table}

In terms of results, Table \ref{table:task_results} shows the performance of the baseline models and our distilled model in the SQuAD-es dataset. The SpanishTinyRoBERTa achieved 80.51\% and 71.23\% for F1 score and Exact Match (EM), respectively, comparable to the performance attained by the teacher model, 87.50\% and 78.30\%, for F1 score and EM, respectively. These findings provide evidence that the SpanishTinyRoBERTa can achieve competitive results on the QA task while requiring much less computational resources. On this regard, we can observe in Appendix \ref{appendix:model_efficiency} that SpanishTinyRoBERTa contains 6.9x less parameters (51M) than the teacher model (355M), and achieves 4.2x inference speedup (392ms vs 1683ms per query). This shows the benefit of using the distillation process to achieve a much lighter, faster and highly performant model.

\section{Conclusion}

In this work we present SpanishTinyRoBERTa, a compressed language model for efficient question answering in Spanish that meets similar performance results to its larger counterpart, with a significant reduction in terms of required computational resources. We showed that the model has the potential to contribute to the adoption of language model for Spanish question answering language-related tasks on resource-constrained environments, while preserving considerable levels of accuracy and robustness. As future work, we aim to produce compressed models for other NLP downstream tasks.

\subsubsection*{URM Statement}

The authors acknowledge that at least one key author of this work meets the URM criteria of ICLR 2024 Tiny Papers Track.

\bibliography{iclr2023_conference_tinypaper}

\begin{thebibliography}{13}
\providecommand{\natexlab}[1]{#1}
\providecommand{\url}[1]{\texttt{#1}}
\expandafter\ifx\csname urlstyle\endcsname\relax
  \providecommand{\doi}[1]{doi: #1}\else
  \providecommand{\doi}{doi: \begingroup \urlstyle{rm}\Url}\fi

\bibitem[Boreshban et~al.(2021)Boreshban, Mirbostani, Ghassem-Sani, Mirroshandel, and Amiriparian]{boreshban2021improving}
Yasaman Boreshban, Seyed~Morteza Mirbostani, Gholamreza Ghassem-Sani, Seyed~Abolghasem Mirroshandel, and Shahin Amiriparian.
\newblock Improving question answering performance using knowledge distillation and active learning, 2021.

\bibitem[Carrino et~al.(2019)Carrino, Costa-jussà, and Fonollosa]{carrino2019automatic}
Casimiro~Pio Carrino, Marta~R. Costa-jussà, and José A.~R. Fonollosa.
\newblock Automatic spanish translation of the squad dataset for multilingual question answering, 2019.

\bibitem[Cañete et~al.(2020)Cañete, Chaperon, Fuentes, Ho, Kang, and Pérez]{CaneteCFP2020}
José Cañete, Gabriel Chaperon, Rodrigo Fuentes, Jou-Hui Ho, Hojin Kang, and Jorge Pérez.
\newblock Spanish pre-trained bert model and evaluation data.
\newblock In \emph{PML4DC at ICLR 2020}, 2020.

\bibitem[de~la Rosa et~al.(2022)de~la Rosa, Ponferrada, Villegas, de~Prado~Salas, Romero, and Grandury]{delarosa2022bertin}
Javier de~la Rosa, Eduardo~G. Ponferrada, Paulo Villegas, Pablo~Gonzalez de~Prado~Salas, Manu Romero, and Marıa Grandury.
\newblock Bertin: Efficient pre-training of a spanish language model using perplexity sampling, 2022.

\bibitem[Devlin et~al.(2019)Devlin, Chang, Lee, and Toutanova]{devlin2019bert}
Jacob Devlin, Ming-Wei Chang, Kenton Lee, and Kristina Toutanova.
\newblock Bert: Pre-training of deep bidirectional transformers for language understanding, 2019.

\bibitem[Gutiérrez-Fandiño et~al.(2022)Gutiérrez-Fandiño, Armengol-Estapé, Pàmies, Llop-Palao, Silveira-Ocampo, Carrino, Armentano-Oller, Rodriguez-Penagos, Gonzalez-Agirre, and Villegas]{gutierrezfandino2022}
Asier Gutiérrez-Fandiño, Jordi Armengol-Estapé, Marc Pàmies, Joan Llop-Palao, Joaquin Silveira-Ocampo, Casimiro~Pio Carrino, Carme Armentano-Oller, Carlos Rodriguez-Penagos, Aitor Gonzalez-Agirre, and Marta Villegas.
\newblock Maria: Spanish language models.
\newblock \emph{Procesamiento del Lenguaje Natural}, 68\penalty0 (0):\penalty0 39--60, 2022.
\newblock ISSN 1989-7553.
\newblock URL \url{http://journal.sepln.org/sepln/ojs/ojs/index.php/pln/article/view/6405}.

\bibitem[Jiao et~al.(2020)Jiao, Yin, Shang, Jiang, Chen, Li, Wang, and Liu]{jiao-etal-2020-tinybert}
Xiaoqi Jiao, Yichun Yin, Lifeng Shang, Xin Jiang, Xiao Chen, Linlin Li, Fang Wang, and Qun Liu.
\newblock {T}iny{BERT}: Distilling {BERT} for natural language understanding.
\newblock In Trevor Cohn, Yulan He, and Yang Liu (eds.), \emph{Findings of the Association for Computational Linguistics: EMNLP 2020}, pp.\  4163--4174, Online, November 2020. Association for Computational Linguistics.
\newblock \doi{10.18653/v1/2020.findings-emnlp.372}.
\newblock URL \url{https://aclanthology.org/2020.findings-emnlp.372}.

\bibitem[Liu et~al.(2019)Liu, Ott, Goyal, Du, Joshi, Chen, Levy, Lewis, Zettlemoyer, and Stoyanov]{liu2019roberta}
Yinhan Liu, Myle Ott, Naman Goyal, Jingfei Du, Mandar Joshi, Danqi Chen, Omer Levy, Mike Lewis, Luke Zettlemoyer, and Veselin Stoyanov.
\newblock Roberta: A robustly optimized bert pretraining approach, 2019.

\bibitem[P{\'e}rez et~al.(2022)P{\'e}rez, Furman, Alonso~Alemany, and Luque]{perez-etal-2022-robertuito}
Juan~Manuel P{\'e}rez, Dami{\'a}n~Ariel Furman, Laura Alonso~Alemany, and Franco~M. Luque.
\newblock {R}o{BERT}uito: a pre-trained language model for social media text in {S}panish.
\newblock In Nicoletta Calzolari, Fr{\'e}d{\'e}ric B{\'e}chet, Philippe Blache, Khalid Choukri, Christopher Cieri, Thierry Declerck, Sara Goggi, Hitoshi Isahara, Bente Maegaard, Joseph Mariani, H{\'e}l{\`e}ne Mazo, Jan Odijk, and Stelios Piperidis (eds.), \emph{Proceedings of the Thirteenth Language Resources and Evaluation Conference}, pp.\  7235--7243, Marseille, France, June 2022. European Language Resources Association.
\newblock URL \url{https://aclanthology.org/2022.lrec-1.785}.

\bibitem[Rajpurkar et~al.(2016)Rajpurkar, Zhang, Lopyrev, and Liang]{rajpurkar2016squad}
Pranav Rajpurkar, Jian Zhang, Konstantin Lopyrev, and Percy Liang.
\newblock Squad: 100,000+ questions for machine comprehension of text, 2016.

\bibitem[Vaswani et~al.(2017)Vaswani, Shazeer, Parmar, Uszkoreit, Jones, Gomez, Kaiser, and Polosukhin]{vaswani2017}
Ashish Vaswani, Noam Shazeer, Niki Parmar, Jakob Uszkoreit, Llion Jones, Aidan~N Gomez, \L~ukasz Kaiser, and Illia Polosukhin.
\newblock Attention is all you need.
\newblock In I.~Guyon, U.~Von Luxburg, S.~Bengio, H.~Wallach, R.~Fergus, S.~Vishwanathan, and R.~Garnett (eds.), \emph{Advances in Neural Information Processing Systems}, volume~30. Curran Associates, Inc., 2017.

\bibitem[Wolf et~al.(2020)Wolf, Debut, Sanh, Chaumond, Delangue, Moi, Cistac, Rault, Louf, Funtowicz, Davison, Shleifer, von Platen, Ma, Jernite, Plu, Xu, Scao, Gugger, Drame, Lhoest, and Rush]{wolf2020huggingfaces}
Thomas Wolf, Lysandre Debut, Victor Sanh, Julien Chaumond, Clement Delangue, Anthony Moi, Pierric Cistac, Tim Rault, Rémi Louf, Morgan Funtowicz, Joe Davison, Sam Shleifer, Patrick von Platen, Clara Ma, Yacine Jernite, Julien Plu, Canwen Xu, Teven~Le Scao, Sylvain Gugger, Mariama Drame, Quentin Lhoest, and Alexander~M. Rush.
\newblock Huggingface's transformers: State-of-the-art natural language processing, 2020.

\bibitem[Yang et~al.(2019)Yang, Dai, Yang, Carbonell, Salakhutdinov, and Le]{yang2019}
Zhilin Yang, Zihang Dai, Yiming Yang, Jaime Carbonell, Russ~R Salakhutdinov, and Quoc~V Le.
\newblock Xlnet: Generalized autoregressive pretraining for language understanding.
\newblock In H.~Wallach, H.~Larochelle, A.~Beygelzimer, F.~d\textquotesingle Alch\'{e}-Buc, E.~Fox, and R.~Garnett (eds.), \emph{Advances in Neural Information Processing Systems}, volume~32. Curran Associates, Inc., 2019.

\end{thebibliography}
\bibliographystyle{iclr2023_conference_tinypaper}

\newpage
\appendix
\section{Appendix}

\subsection{Details on the distillation process} \label{appendix:distillation_process}

In this section, we provide further details on the training distillation process. As mentioned in Section \ref{sec:methodology}, we utilize the knowledge distillation process described in \citep{jiao-etal-2020-tinybert} with an addition term to account for task-specific loss. More specifically, we assume that the teacher and student models have $K$ and $L$ Transformer layers, respectively, where $L$ $\ll$ $K$. To account for the discrepancy in terms of number of layers for the student and teacher models, we use a layer index mapping function, $g(l)$ = 3 x $k$, such that the student learns from every 3 layers of the teacher model. Therefore, the student acquires knowledge from the teacher by minimizing the following loss function:

\begin{align}
\!\!\!\mathcal{L} \! = \! \sum_{x \in \mathcal{X}} \mathcal{L}_{\text{task}}(f^S(x), f^T(x)) + \sum^{K+1}_{k=0} \mathcal{L}_{\text{layer}}(f^S_k(x), f^T_{g(k)}(x)),
\end{align}

where $\mathcal{L}_{\text{layer}}$ is the loss function of a given model layer (e.g. a Transformer or embedding layer), $f_k(x)$ denotes the behavior function associated with the $k$-th layer of the model and $\mathcal{L}_{\text{task}}$ is the task-specific distillation loss function applied to the student and teacher models outputs. The layer loss, $\mathcal{L}_{\text{layer}}$, is calculated as the sum of two terms, $\mathcal{L}_{\text{attention}}$ and $\mathcal{L}_{\text{hidden}}$, for the attention scores and hidden representations, respectively. Specifically, $\mathcal{L}_{\text{attention}}$ is modelled as:

\begin{align}
\mathcal{L}_{\text{attention}} = \frac{1}{h}\sum\nolimits^{h}_{i=1} \texttt{MSE}(\bm{A}_i^{S}, \bm{A}_i^{T}),
\end{align}

where $h$ is the number of attention heads, $\bm{A}_i \in \mathbb{R}^{l\times l} $ is the $i$-th head attention matrix of teacher or student, $l$ is the input text length and {\tt MSE()} depicts the mean squared error loss function. On other side, $\mathcal{L}_{\text{hidden}}$ is defined as:

\begin{align}
\mathcal{L}_{\text{hidden}} = \texttt{MSE}(\bm{H}^{S}\bm{W}_h, \bm{H}^{T}), 
\end{align}

where the matrices $\bm{H}^{S} \in \mathbb{R}^{l\times d'}$ and $\bm{H}^{T} \in \mathbb{R}^{l \times d}$ refer to the hidden states of student and teacher networks respectively. The scalar values $d$ and $d'$ denote the hidden sizes of teacher and student models, and $d'$ is often smaller than $d$ to obtain a smaller student network. The matrix $\bm{W}_h \in \mathbb{R}^{d' \times d} $ is a learnable linear transformation, which projects the hidden states of the student network into the same dimensionality as the teacher feature space. Therefore, $\mathcal{L}_{\text{layer}}$ is expressed as:

\begin{align}
\mathcal{L}_{\text{layer}} = \mathcal{L}_{\text{attention}} + \mathcal{L}_{\text{hidden}}
\end{align}

By following this distillation process, the smaller model is able to mimic the representation and attention of the teacher over the input sequences, while maximizing the downstream task performance.

\subsection{Details on training hyperparameters and environment} \label{appendix:training_details}

Training was done using a single NVIDIA RTX A6000 GPU, with a batch size of 32, a learning rate of 3e-5, and maximum sequence length of 384 running for 20 epochs. For optimization purposes, mixed-precision training is employed, and gradient clipping with a max gradient norm of 1.0 is applied to improve training stability. The distillation process took 5.3 hours to complete. We utilize the HuggingFace library \citep{wolf2020huggingfaces} for our training implementation and the source code is available at \href{https://github.com/anonymous/anonymous}{https://github.com/anonymous/anonymous}.

\subsection{Experimental results on model efficiency} \label{appendix:model_efficiency}

In Table \ref{table:model_statistics} we show a comparison of model sizes and inference latency between the baseline models and our SpanishTinyRoBERTa model. The results were obtained by running on a single NVIDIA RTX A6000 GPU and averaging over 10 different runs.

\begin{table}[H]
    \renewcommand\arraystretch{1.0}
    \centering
    \small
\begin{tabular}{lccccc}
\toprule
Model & Layers & Hidden Size & Feed-forward Size & Size & Latency (ms) \\
\hline
Spanish RoBERTa-large (teacher)  & 24 & 1024 & 4096 & 355M & 1683     \\ 
Multilingual BERT (mBERT) & 12 & 768 & 3072 & 179M & 1187 \\ 
Spanish BERT-base & 12 & 768 & 3072 & 109M & 942 \\ 
Spanish RoBERTa-base & 12 & 768 & 3072 & 124M & 1015   \\ 
\hline
SpanishTinyRoBERTa (ours) & 6 & 512 & 3072 & 51M & 392  \\ 
\hline
\end{tabular}
  \caption{Comparison of model sizes and inference latency (in milliseconds for a single query) between the baselines and our distilled model. The number of layers does not include the embedding and prediction layers.}
  \label{table:model_statistics}
\end{table}

\end{document}